\documentclass[letterpaper, 10 pt, conference]{ieeeconf}  

\IEEEoverridecommandlockouts                              

\usepackage{graphics} 
\usepackage{epsfig} 
\usepackage{mathptmx} 
\usepackage{times} 
\usepackage{amsmath} 
\usepackage{amssymb}  
\usepackage{subfigure}
\usepackage{multirow}
\usepackage{graphicx, epstopdf}
\usepackage{xspace,epsfig,url}
\usepackage[noadjust]{cite}
\usepackage{enumerate}
\usepackage{float}
\usepackage{epsf}
\usepackage{psfrag}
\usepackage{verbatim}
\usepackage[all]{xy}
\usepackage{color}
\usepackage[table]{xcolor}
\usepackage{cases}
\usepackage{hyperref}

\title{\LARGE \bf
Human-centered Control of a Growing Soft Robot for Object Manipulation*
}

\author{Fabio Stroppa, Ming Luo, Giada Gerboni, Margaret M. Coad, Julie M. Walker, and Allison M. Okamura
\thanks{*Toyota Research Institute (``TRI'')  provided funds to assist the authors with their research but this article solely reflects the opinions and conclusions of its authors and not TRI or any other Toyota entity. The authors thank A. Thackston and S. Zapolsky for their ideas related to this work.}
\thanks{The authors are with the Mechanical Engineering Department, Stanford University, Stanford, CA 94305, USA
        {\tt\small fstroppa@stanford.edu}}%
}

\begin{document}

\maketitle
\thispagestyle{empty}
\pagestyle{empty}

\begin{abstract}

We present a user-friendly interface to teleoperate a soft robot manipulator in a complex environment. Key components of the system include a manipulator with a grasping end-effector that grows via tip eversion, gesture-based control, and haptic display to the operator for feedback and guidance. In the initial work, the operator uses the soft robot to build a tower of blocks, and future works will extend this to shared autonomy scenarios in which the human operator and robot intelligence are both necessary for task completion. 
\end{abstract}

\section{INTRODUCTION}

Robots in domestic environments have the potential to both autonomously assist humans and enable a physical presence for remote operators. 
A variety of operation modalities can potentially lie in the spectrum between autonomy and teleoperation, providing a rich and useful set of robot capabilities. However, this also creates numerous challenges for robot design and intuitive human-in-the-loop control.
Whether supervising an autonomous robot or directly controlling a teleoperated one, human operators require intuitive interfaces and a sense of immersion while dealing with a remote environment.
Haptic feedback can improve intuition and immersion by directly communicating physical interactions and constraints, and there is the opportunity to design haptic interfaces with the specific feedback modalities and degrees of freedom appropriate for shared autonomy scenarios.


In this work, we present a setup including a user-friendly Motion Capture (MoCap) system to teleoperate a growing soft robot during a manipulation task.
The MoCap allows the operator to interact with the robot using an intuitive approach, removing the burden of learning the mapping required by a joystick or another physical interface. 
Furthermore, the operator is also equipped with a light holdable controller providing kinesthetic and tactile haptics for feedback and guidance, facilitating manipulation in complex and unstructured environments.

\begin{figure}[t!]
  \centering
	{\includegraphics[width=\linewidth]{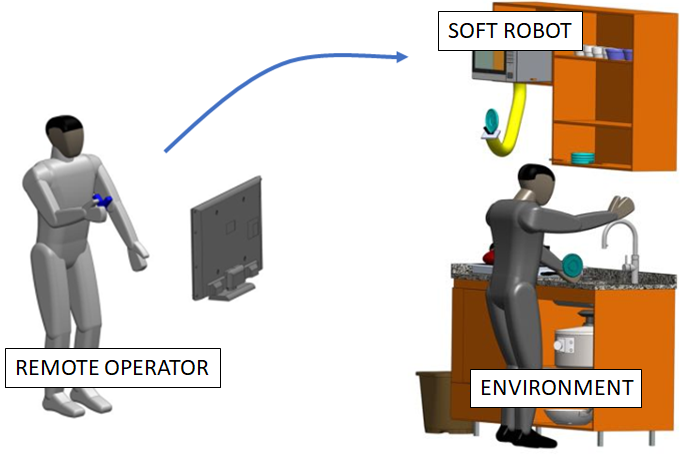}}
    \caption{Concept of teleoperation and shared autonomy with a soft robot in a human environment.}
    \label{fig:system}
		\vspace{-0.6cm}
\end{figure}

Although the scenario considered here consists of simple reaching, grasping, and manipulation of blocks, there are many other applications for shared control of robots in homes or other complex environments.
Fig.~\ref{fig:system} shows a possible example, in which a remote human operator uses a soft robot to help other humans in a kitchen. In particular, the operator may have a direct line of sight of the robot and being physically in front of it, or can be interfaced with a visual interface (e.g. a screen or a head mounted display) for a remote teleoperation.

\section{BACKGROUND}

Robot manipulators are desirable for assistance with tasks in domestic environments, although their safety is often limited by their inherent mechanical properties.
Soft continuum robots offer the potential for safe physical interactions with humans, and exhibit access and manipulation capabilities in constrained and cluttered environments not achievable by traditional robots \cite{trivedi2008soft}. 
However, environmental contact can drastically alter the motion of soft robots, complicating their control and limiting interaction forces \cite{yip2014model}.
Prior studies of the authors introduced a novel soft robot that extends by growing from its tip and controls the direction of growth by reversible bending \cite{greer2017series}. This robot has been used to navigate cluttered environments and steer toward targets \cite{coad2019vine}, but its use in manipulation has not yet been studied. 

When the manipulation task is performed in a shared-autonomy scenario, and the operator can exert control over some aspects of the task, sharing the control of the object handling forces is the more natural approach, and it also allows the human operator to receive force information as feedback \cite{griffin2005feedback}. 
Artificial haptic sensations can present information to operators
, but the majority of existing devices have focused on applying forces to the operators’ finger pads or recreating  textures \cite{benko2016normaltouch, whitmire2018haptic}. 
A previous study from the authors proposed a device that generates force and torque sensations by applying tangential cues to the finger pads \cite{walker2019holdable}, specifically for motion guidance. This device is used in the current setup as part of the interface for sharing the autonomy between the operator and the robot.




\section{MATERIALS AND METHODS}

\begin{figure}[t!]
  \centering
	
	{\includegraphics[width=\linewidth]{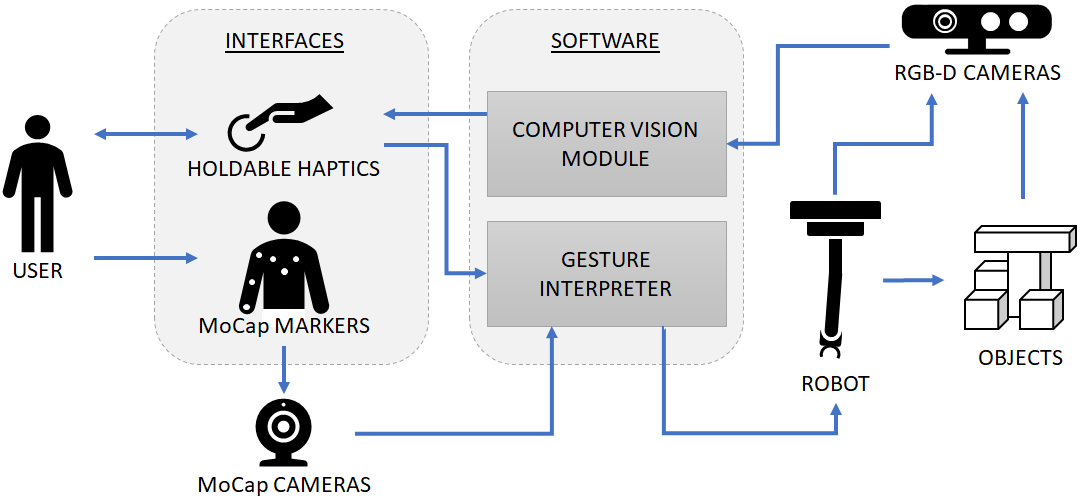}}
    \caption{System components for human-centered control of a soft robot.}
    \label{fig:schema}
	  \vspace{-0.5cm}
\end{figure}

The proposed task consists of manipulating toy blocks in order to build a ``block tower'' by stacking them on top of one another. The robot is attached on the ceiling and grasps the blocks from above. Fig.~\ref{fig:schema} shows the proposed system, composed of the soft robot, two main interfaces, two high-level software modules, and a set of cameras for both the MoCap system and the 3D tracking.

\begin{figure}[t!]
\centering
	\subfigure[\label{fig:vine_robot}]%
	{\includegraphics[width=7cm]{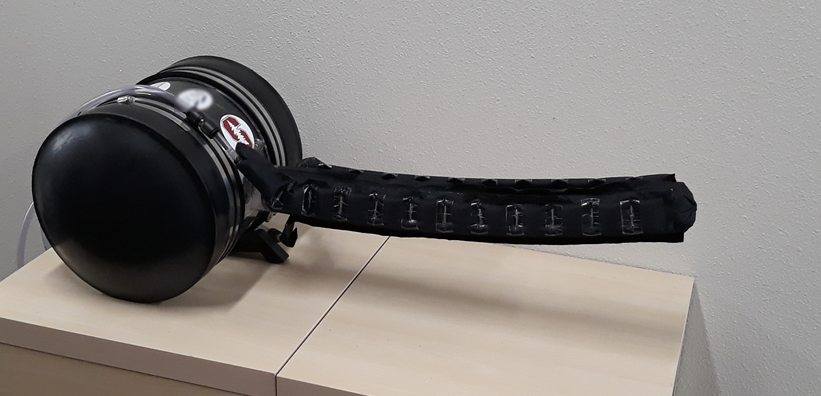}}\qquad
	\subfigure[\label{fig:gripper}]%
	{\includegraphics[height=3cm]{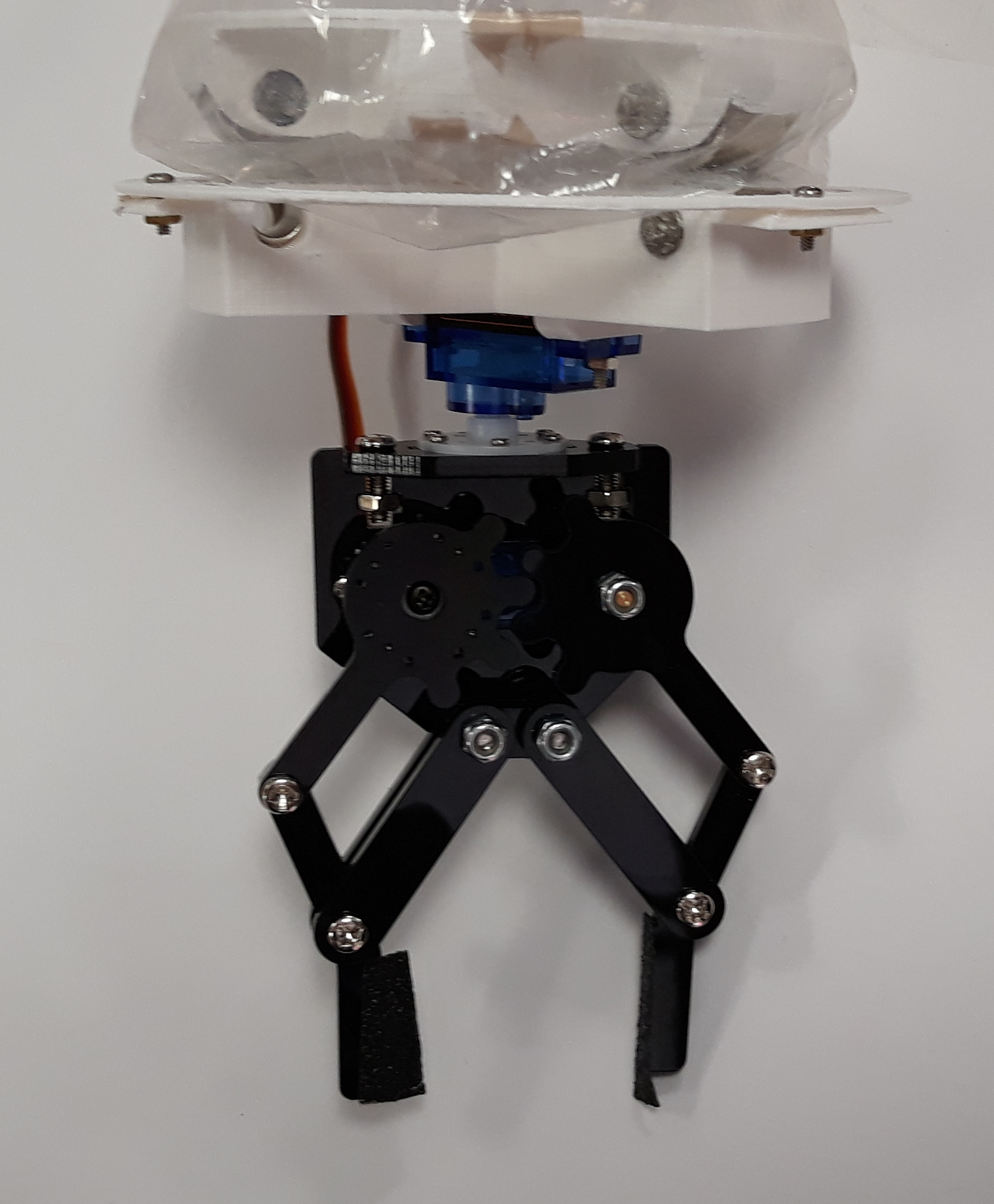}}\qquad	
	\subfigure[\label{fig:holdable_haptics}]%
	{\includegraphics[height=3cm]{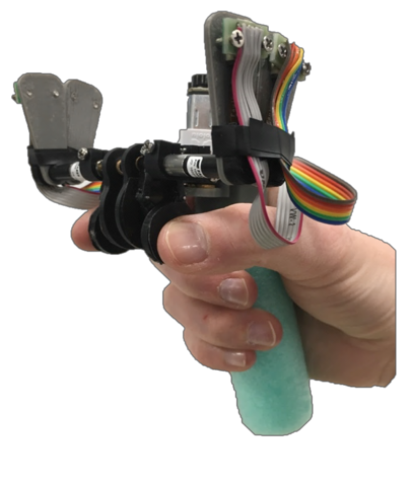}}
	\caption{Existing components of the human-centered soft robot manipulation system: (a) a steerable soft robot, (b) a one-degree-of-freedom gripper, and (c) a four-degree-of-freedom holdable device for haptic feedback.}\label{fig:methods}
	\vspace{-0.6cm}
\end{figure}

\textbf{Robot:} The soft robot (Fig.~\ref{fig:vine_robot}) can grow via pneumatic actuation to an arbitrary length and steer to reach the blocks in the workspace. 
Once a block is reached, it can be grasped and handled by the gripper equipped at the tip of the robot (see Fig.~\ref{fig:gripper}). 
The base of the gripper is composed of two rings, one inside and one outside the robot, which are connected by rolling magnets. The material of the soft robot can slide through the surface between the magnets when growing or retracting, such that the gripper remains stable on the tip position. 

\textbf{Interfaces:}
The MoCap system is used to control the robot. The PhaseSpace Impulse X2E (phasespace.com) was chosen to track the movements of the operator.  
Commands are sampled at $270~Hz$ and sent to the robot at $10~Hz$.

The haptic device is an holdable interface that allows the operator to move around freely in a large workspace. It consists of a hand-held kinesthetic gripper (Fig.~\ref{fig:holdable_haptics}) that provides guidance cues in $4$ degrees of freedom through skin stretch and kinesthetic feedback at the fingertips. 
A motor at the hinge allows the fingers to open and close to perform grasping movements, which controls the gripper at the tip of the robot.

\textbf{Software:}
A gesture interpreter is used to map the operator's movements to the kinematics of the soft robot. It recognizes multiple commands such as grow/retract, left/right, up/down -- which can be given simultaneously -- and is independent from the MoCap reference frame, such that the operator is not asked to stand in a particular position and is free to move comfortably.

A set of RGB-D cameras (2D image from RGB optical cameras and depth data from infrared cameras) are used for 3D reconstruction of the environment. The robot and the blocks are recognized and tracked in real time ($\approx30~Hz$) by a 3D-processing engine that analyzes their pose in the space, in order to evaluate a strategy for the robot to reach the targets and handle the task. Based on this evaluation, the module sends cues to the haptic device to help the operator navigate the environment, pointing in the direction of the desired target or away from any dangerous regions.

\section{CONCLUSION}

Future works include further development of shared autonomy and teleoperation scenarios, defining metrics for performance, integrating new soft robot and haptic device designs, and creating novel interactions with a variety of perception, modeling, and planning/control strategies.

\addtolength{\textheight}{-12cm}   





\bibliographystyle{IEEEtran}
\bibliography{worldhaptics19_wip}

\end{document}